\documentclass[final]{cvpr}

\usepackage{times}
\usepackage{epsfig}
\usepackage{graphicx}
\usepackage{amsmath}
\usepackage{amssymb}

\usepackage[pagebackref=true,breaklinks=true,colorlinks,bookmarks=false]{hyperref}

\def\cvprPaperID{****} 
\def\confYear{CVPR 2021}

\begin{document}

\title{AttendSeg: A Tiny Attention Condenser Neural Network for Semantic Segmentation on the Edge}

\author{Xiaoyu Wen$^{3,*}$, Mahmoud Famouri$^{3,*}$, Andrew Hryniowski$^{1,3,*}$, Alexander Wong$^{1,2,3,*}$\\
$^1$Department of Systems Design Engineering, University of Waterloo\\
$^2$Waterloo Artificial Intelligence Institute, University of Waterloo\\
$^3$DarwinAI Corp.\\
{\tt\small$^*$equal contribution}

}

\maketitle

\begin{abstract}
In this study, we introduce \textbf{AttendSeg}, a low-precision, highly compact deep neural network tailored for on-device semantic segmentation.  AttendSeg possesses a self-attention network architecture comprising of light-weight attention condensers for improved spatial-channel selective attention at a very low complexity.  The unique macro-architecture and micro-architecture design properties of AttendSeg strike a strong balance between representational power and efficiency, achieved via a machine-driven design exploration strategy tailored specifically for the task at hand.  Experimental results demonstrated that the proposed AttendSeg can achieve segmentation accuracy comparable to much larger deep neural networks with greater complexity while possessing a significantly lower architecture and computational complexity (requiring as much as $>$27$\times$ fewer MACs, $>$72$\times$ fewer parameters, and $>$288$\times$ lower weight memory requirements), making it well-suited for TinyML applications on the edge.
\end{abstract}
\section{Introduction}
\label{introduction}

Semantic segmentation remains a challenging task in computer vision, with the underlying goal being to assign class labels on a per-pixel basis to an image.  Much of the success in semantic segmentation has revolved around deep learning~\cite{lecun2015deep}, where deep convolutional neural networks have been designed to model the relationship between input images and output label fields~\cite{ronneberger2015unet,lin2017refinenet,TuSimple,PSPNet,chen2018deeplab,deeplab3,deeplab3plus2018,Araslanov_2020_CVPR,Li_2020_CVPR,Yang_2020_CVPR}.  Exemplary deep semantic segmentation network architectures include U-Net~\cite{ronneberger2015unet}, RefineNet~\cite{lin2017refinenet}, TuSimple~\cite{TuSimple}, PSPNet~\cite{PSPNet}, and the DeepLab family of networks~\cite{chen2018deeplab,deeplab3,deeplab3plus2018}.  Despite such advancements, a major bottleneck to the widespread adoption of deep semantic segmentation networks in on-device, TinyML applications has been their high architectural and computational complexity.

Given the significant resource constraints imposed by low-cost, low-power edge devices, there has been great recent interest in strategies for producing highly compact networks tailored for low-power, on-device edge usage. These include efficient design principles~\cite{MobileNetv1,MobileNetv2,SqueezeNet,SquishedNets,TinySSD,ShuffleNetv1,ShuffleNetv2,ResNet}, precision reduction~\cite{Jacob,Meng2017,courbariaux2015binaryconnect}, model compression~\cite{han2015deep,hinton2015distilling,projectionnet}, network architecture search~\cite{MONAS,ParetoNASH,MNAS,shafiee2017deep}, dynamic routing~\cite{Li_2020_CVPR}, and efficient self-attention~\cite{wong2020tinyspeech,wong2020attendnets}.

In this study, we introduce AttendSeg, a low-precision, highly compact deep semantic segmentation network tailored for TinyML applications.  By leveraging both machine-driven design exploration and attention condensers, the proposed AttendSeg possesses a highly efficient self-attention architecture with unique macro-architecture and micro-architecture designs tailored specifically for low-power, on-device semantic segmentation.

This paper is organized as follows.  Section 2 describes the underlying methodology behind the design of the proposed AttendSeg network architecture for semantic segmentation on the edge.  Experimental results are presented and discussed in Section 3. Conclusions are drawn and future work are discussed in Section 4.

\begin{figure*}[ht!]
    \centering
        \includegraphics[width=\textwidth]{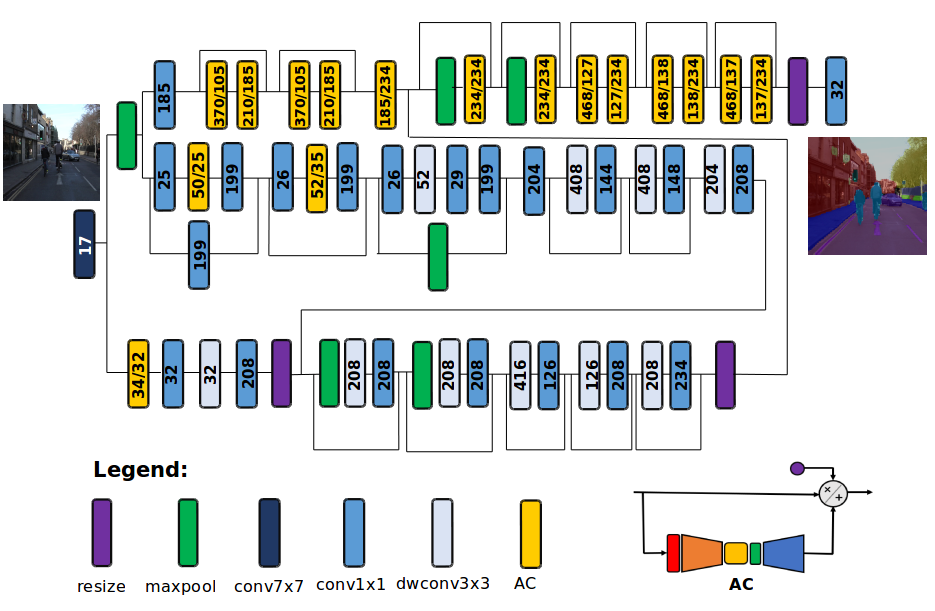}
    \caption{The network architecture of AttendSeg network for semantic segmentation.  The proposed AttendSeg network architecture exhibits heavy use of lightweight attention condensers, depthwise convolutions, and pointwise convolutions, selective long-range connectivity, and aggressive dimensionality reduction via strided convolutions with large strides, and high architectural heterogeneity.  All of these properties of AttendSeg lead to a strong balance between representational power and efficiency.}
    \label{fig:AttendSeg}
\end{figure*}
\section{Methods}
\label{method}
This work presents AttendSeg, a low-precision, highly compact deep neural network architecture is tailored specifically for on-device semantic segmentation.  Two key concepts are leveraged in the construction of the proposed AttendSeg: 1) attention condensers for enabling highly efficient selective attention, and 2) machine-driven design exploration to construct the macro-architecture and micro-architcture designs of the proposed deep neural network architecture.  Both of these concepts, along with the final AttendSeg architectural design, are described in detail below.

\subsection{Attention Condensers}
The first concept we leverage to construct the proposed AttendSeg is the concept of attention condensers~\cite{wong2020tinyspeech,wong2020attendnets}.  One of the breakthroughs in deep learning in recent years has been the concept of self-attention~\cite{vaswani2017attention,hu2017squeezeandexcitation,woo2018cbam,devlin2018bert}, which is inspired by human selective attention that filters out unimportant details to focus on what matters for the task at hand.  Much of literature in self-attention has focused on accuracy, which has heavily influenced the design of such mechanisms.  Recently, an efficient self-attention mechanism was introduced in the form of attention condensers~\cite{wong2020tinyspeech,wong2020attendnets}.  More specifically, attention condensers enable highly efficient selective attention by learning condensed embeddings of activation relationships, and in the case of AttendSeg they serve specifically for capturing joint spatial-channel activation relationships.  An attention condenser consists of a condensation layer to reduce spatial-channel dimensionality, an embedding structure to characterize joint spatial-channel activation relationships, an expansion layer to increase dimensionality, and a selective attention mechanism for imposing selective attention.

\begin{table*}[t]
\centering
\caption{Performance of tested networks on CamVid. Best results in \textbf{bold}.  Results for AttendSeg based
on 8-bit weights, while other tested networks based on 32-bit weights.}
\begin{tabular}{c|c|c|c|c}
~~~~~~~Model~~~~~~~                                  & ~~Accuracy (\%)~~             & MACs (G) & Parameters (M) & Weight Memory (Mb)         \\ \hline
\hline
\multicolumn{1}{c|}{RefineNet~\cite{lin2017refinenet}} & \textbf{90.0\% }         & 202.47         & 85.69   & 343 \\
\multicolumn{1}{c|}{EdgeSegNet~\cite{lin2019edgesegnet}}         & 89.15\%          & 77.89 & 7.09 & 28.3\\
\multicolumn{1}{c|}{AttendSeg}         & 89.89\%          & \textbf{7.45} & \textbf{1.19} & \textbf{1.19}
\end{tabular}
\label{tab_Results}
\end{table*}

\subsection{Machine-driven design exploration}
To construct the macro-architecture and micro-architecture designs of AttendSeg, we leverage a machine-driven design exploration strategy tailored around the operational requirements of on-device semantic segmentation.  Here, we harness the concept of generative synthesis~\cite{Wong2018}, where the design exploration problem is framed as a constrained optimization problem to identify a generator $\mathcal{G}$ that, given a set of seeds $S$, can generate networks $\left\{N_s|s \in S\right\}$ maximizing a universal performance function $\mathcal{U}$ (e.g.,~\cite{Wong2018_Netscore}) while satisfying requirements defined via an indicator function $1_r(\cdot)$,
{\small
\begin{equation}
\mathcal{G}  = \max_{\mathcal{G}}~\mathcal{U}(\mathcal{G}(s))~~\textrm{subject~to}~~1_r(\mathcal{G}(s))=1,~~\forall s \in S.
\label{optimization}
\end{equation}
}
An approximate solution $\hat {\mathcal{G}}$ to  Eq.~\ref{optimization} can be obtained via iterative optimization, initialized based on a design prototype $\varphi$, $\mathcal{U}$, and $1_r(\cdot)$.  The resulting $\hat{\mathcal{G}}$ can be thus used to generate the final AttendSeg architecture design.  Here, the design prototype $\varphi$ is based on the multi-path refinement design principles introduced in~\cite{lin2017refinenet}, which facilitate the refinement of high-level semantic representation in deeper layers based on fine-grained representation in earlier layers.  The indicator function $1_r(\cdot)$ is defined such that: 1) an accuracy of $\geq$ 88\% is achieved on CamVid~\cite{CamVid}, to be within 2\% of EdgeSegNet~\cite{lin2019edgesegnet}, a state-of-the-art efficient deep semantic segmentation network, and 2) 8-bit weight precision.

\subsection{Architectural Design}
\label{design}

The network architecture of the AttendSeg is shown in Fig.~\ref{fig:AttendSeg}, which possesses several interesting properties.  First, AttendSeg is comprised of a heterogeneous mix of lightweight attention condensers, depthwise convolutions, and pointwise convolutions with unique micro-architecture designs, thus striking a strong balance between representational power and efficiency.  Second, AttendSeg exhibits selective long-range connectivity where only select deeper layers are refined based on earlier layers, thus improving architectural efficiency by only refining at scales that benefit from it.  Third, one can observe very aggressive dimensionality reduction via strided convolutions with large strides, thus greatly reducing complexity while preserving representational capacity.  These properties illustrate the power of leveraging both machine-driven design exploration and attention condensers to produce highly compact network architectures tailored for edge scenarios.

\section{Results and Discussion}
\label{results}
We explore the efficacy of AttendSeg for on-device semantic segmentation on the edge using the Cambridge-driving Labeled Video Database (CamVid)~\cite{CamVid}, a dataset introduced for evaluating semantic segmentation performance with 32 different semantic classes. For comparison purposes, the results for ResNet-101 RefineNet~\cite{lin2017refinenet} and EdgeSegNet~\cite{lin2019edgesegnet}, a state-of-the-art efficient deep semantic segmentation network, are also presented and all experiments are conducted at a 512$\times$512 resolution in TensorFlow.

It can be seen in Table~\ref{tab_Results} that the proposed AttendSeg achieved similar accuracy as ResNet-101 RefineNet and higher than EdgeSegNet, but has $>$\textbf{72$\times$} and $>$\textbf{5.9$\times$} fewer parameters compared to RefineNet and EdgeSegNet, respectively.  Due to the low-precision nature of AttendSeg, its weight memory requirements are $>$\textbf{288$\times$} and $>$\textbf{23.6$\times$} lower than RefineNet and EdgeSegNet, respectively.  More interestingly, AttendSeg achieves $>$\textbf{27$\times$} and $>$\textbf{10.4$\times$} greater computational efficiency in terms of multiply-accumulate (MAC) operations compared to RefineNet and EdgeSegNet, respectively.  An example semantic segmentation outputs produced using AttendSeg is shown in Fig.~\ref{fig:Camvid}.  Visually, it can be observed that AttendSeg achieves good segmentation performance.
{
\begin{figure}[h]
\centering
    \includegraphics[width=0.4\textwidth]{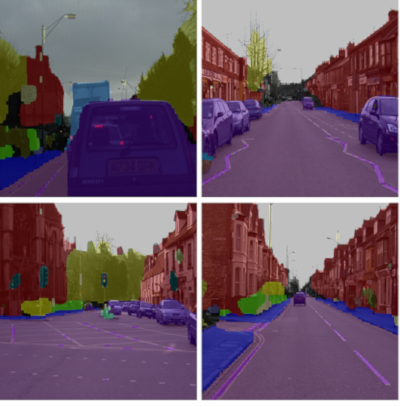}
    \caption{Example semantic segmentations using AttendSeg.}
    \label{fig:Camvid}
\end{figure}
}
These experimental results demonstrate that AttendSeg can achieve  strong semantic segmentation performance while possessing significantly lower architectural and computational complexity, making it well-suited for TinyML applications where resources are limited such as mobile, drone, vehicle, and robotic applications.

\section{Conclusion}
In this work, we introduced AttendSeg, a low-precision, highly efficient self-attention deep neural network architecture tailored for semantic segmentation on the edge.  By leveraging a machine-driven design exploration strategy to discover unique macro-architecture and micro-architecture designs as well as leveraging the concept of attention condensers, the proposed AttendSeg architecture strikes a strong balance between representational power and efficiency.  Experimental results show the efficacy of AttendSeg in achieving comparable segmentation accuracy with significantly more complex deep neural network architectures while achieving significantly lower architectural complexity, lower computational complexity, and lower weight memory requirements, making it very well-suited for TinyML applications on edge and embedded devices.  Future work involves exploring this approach for tackling other complex visual perception tasks such as object detection, instance segmentation, and depth estimation.

\section{Broader Impact}

TinyML (tiny machine learning) has been seen significant rise in attention  in recent years as a disruptive technology that will accelerate the widespread adoption of machine learning across industries and society.  In particular, the ability to perform real-time predictions automatically using machine learning on low-cost, low-power edge and embedded devices can enable a huge swath of applications ranging from autonomous vehicles and advanced driving assistance systems~\cite{Bagloee} to intelligent exoskeletons leveraging embedded sensing information for environmental-adaptive control~\cite{Laschowski2020,Laschowski2021}.  In addition, TinyML can enable greater privacy in machine learning applications by facilitating for tetherless intelligence without the need for continuous connectivity or at the very least reduce the amount of information that needs to be sent to the cloud.  The hope is that knowledge and insights gained from TinyML research such as AttendSeg can contribute to the advancement of efforts in TinyML towards ubiquitous machine learning.

Despite all of these potential advantages and benefits of TinyML, it is also important to keep additional considerations in mind with regards to the design and development of such TinyML advancements in terms of not only potential sources of error and biases, but also take socioeconomical considerations into account to better understand how the availability of such technologies can impact society (e.g., privacy, inclusion, ethics, human-machine interaction, new risks, etc.)~\cite{Hancock7684,wang2019implications,cunneen,Cunneen2}.

{\small
\bibliographystyle{ieee_fullname}
\bibliography{egbib}
}

\end{document}